\documentclass{article}
\usepackage{amsmath}
\usepackage{graphicx}
\usepackage{hyperref}
\usepackage{float}
\usepackage{amsfonts}
\usepackage{amssymb}
\usepackage{booktabs}
\usepackage{geometry}
\usepackage[affil-it]{authblk}
\title{Ozone level forecasting in Mexico City with temporal features and interactions}

\author[1$\ast$]{J. M. S\'anchez Cerritos (jmsc@xanum.uam.mx)}
\author[1]{J. A. Mart\'inez-Cadena (martinezcadenajuan@gmail.com)}
\author[2]{A. Mar\'in-L\'opez (marinlopabi@gmail.com)}
\author[1]{J. Delgado-Fern\'andez (jdf@xanum.uam.mx)}
\affil[1]{Departamento de Matemáticas. Universidad Autónoma Metropolitana-Iztapalapa, Iztapalapa, CDMX, 09340 México.}
\affil[2]{Departamento de Ingeniería de Procesos e Hidráulica. Universidad Autónoma Metropolitana-Iztapalapa, Iztapalapa, CDMX, 09340 México.}

\date{}

\begin{document}

\maketitle
$^*${\bf Corresponding Author}: jmsc@xanum.uam.mx

\begin{abstract}
Tropospheric ozone is an atmospheric pollutant that negatively impacts human health and the environment. Precise estimation of ozone levels is essential for preventive measures and mitigating its effects. This work compares the accuracy of multiple regression models in forecasting ozone levels in Mexico City,  first without adding temporal features and interactions, and then with these features included. Our findings show that incorporating temporal features and interactions improves the accuracy of the models.
\end{abstract}

\section{Introduction}

A type of secondary pollution known as tropospheric ozone (O$_3$) is created when precursors such as nitrogen oxides (NO$_x$) and volatile organic compounds (VOCs) react chemically in the presence of sunlight. Through the absorption of UV light, \(NO_2\) molecules are broken down in these photochemical reactions, releasing oxygen atoms that mix with other oxygen molecules (\(O_2\)) to produce ozone (\(O_3\)). The following chemical reactions can summarize this process:

\begin{align*}
    NO_2 + h\nu &\rightarrow NO + O \\
    O + O_2 &\rightarrow O_3 \\
    NO + O_3 &\rightarrow NO_2 + O_2
\end{align*}

Precursor concentration and solar radiation intensity determine the dynamic equilibrium between ozone creation and destruction. Tropospheric ozone is a dangerous pollutant that can lead to a number of health problems as well as environmental difficulties. In contrast, stratospheric ozone creates a protective ozone layer.

Exposure to high levels of tropospheric ozone can cause a range of respiratory problems, including coughing, throat irritation, and worsening of asthma symptoms. Long-term exposure can lead to more severe health issues such as chronic respiratory diseases, reduced lung function, and increased mortality rates. Children, the elderly, and individuals with pre-existing health conditions are particularly vulnerable to the adverse effects of ozone.

Ground-level ozone can also damage flora, which can result in decreased agricultural production, damage to forests, and a decline in biodiversity. It prevents plants from photosynthesizing, which slows down their growth and increases their vulnerability to pests, illnesses, and harsh weather condition. This may have a domino effect on ecosystems, changing the availability of food and the standard of habitat for different species \cite{EPA,WHO}.

Tropospheric ozone is not emitted directly into the atmosphere. Instead,  it originates from photochemical interactions between NO$_x$ and VOCs , which are released from a variety of sources such as vehicles, factories, and natural sources like wildfires.  When these precursors are exposed to sunlight, they undergo complex chemical reactions resulting in ozone.

Monitoring and regulating tropospheric ozone levels is challenging due to the complexity of its formation and the wide range of contributing sources. Urban areas often experience higher levels of ozone due to the concentration of emission sources and favorable conditions for photochemical reactions.  Coordination of efforts to limit precursor emissions, precise forecasting models, and extensive monitoring are necessary for efficient control.

To safeguard the environment and human health in a timely manner, accurate ozone level forecasting is crucial.  Early warnings can help vulnerable populations take protective actions, and policymakers can devise strategies to reduce emissions and mitigate ozone formation. Better forecasting models can help us understand the dynamics of ozone and provide support from more effective regulatory policies.

This study focuses on improving the forecasting accuracy of ozone levels in Mexico City, a metropolitan area that frequently faces high ozone concentrations. We evaluate the effectiveness of several regression models, first without temporal features and interactions, and then with these features included. Our goal is to show the significance of temporal context and variable interactions in enhancing model performance.

\section{Data}

The data used in this study include information on Mexico City's air quality and cover the time period from January 1, 2015 to May 31, 2023.  Air pollutants are monitored by the Red Automatica de Monitoreo Ambiental (RAMA),and meteorological data are provided by Red de Meteorología y Radiación Solar (REDMET), through 28 stations. The website www.aire.cdmx.gob.mx/, which is open to the public, provided the ozone pollution data (3073 daily-averaged observations).  Measurement gaps were filled using the Diosdado et al. (2013) approach. The percentage of gaps was kept below 0.5\%, and the maximum number of consecutive gaps was kept below 12 registers. The variables included are:

\begin{itemize}
    \item \textbf{Carbon Monoxide (CO):} A toxic gas primarily emitted by motor vehicles and industrial processes.
    \item \textbf{Nitrogen Oxides (NO$_X$,  NO and NO$_2$):} Key pollutants that contribute to the formation of tropospheric ozone.
    \item \textbf{Particulate Matter (PM$_{10}$ and PM$_{2.5}$):} Particles suspended in the air that can have adverse effects on respiratory health.
    \item \textbf{Sulfur Dioxide (SO$_2$):} A pollutant from the burning of fossil fuels that can contribute to the formation of fine particles.
    \item \textbf{Ozone (O$_3$):} The target pollutant of this study, whose concentration is to be predicted.
    \item  \textbf{Relative humidity (RH)}.
    \item \textbf{Temperature (TMP)}.
    \item \textbf{Wind direction (WDR)}.
    \item \textbf{Wind speed (WSP)}.
\end{itemize}

Table \ref{sta} presents summary statistics for the variables. Ten of these, show positive skewness, indicating a trend toward extremely high values for all twelve variables rather than a trend toward low values. All variables (except for RH) have thick tails, as shown in terms of kurtosis and Shapiro-Wilk statistics.

\begin{table}
\begin{tabular}{lccccccc}
\toprule
Pollutant &	Min  & Mean  & 	Max & SD & 	Skewness & 	Kurtosis &	Shapiro-Wilk\\
 & ($\mu g/m^3$)	 &  ($\mu g/m^3$) &  ($\mu g/m^3$)	&  ($\mu g/m^3$) &  & 	 &	 \\

\midrule
CO & 	0.116	& 0.481	 & 1.831	& 0.205	& 1.063	& 1.683 & 	0.938(0.00)\\
NO	 & 1.168	& 14.712	& 75.806	 &  8.857	 & 1.265	&  2.462	& 0.914(0.00)\\
NO2	 & 8.061	& 23.030	& 72.193	 & 6.754	 & 0.773	&  1.428	& 0.970(0.00)\\
NOX	 & 9.983	& 37.996	& 148.005	 & 14.795	 & 0.977	&  1.776	& 0.950(0.00)\\
O3 &	3.226 &	30.486 &	67.443 &	9.023	& 0.374	& 0.038 & 	0.991(0.00)\\
PM$_{10}$ & 	8.218 & 	41.892	 & 115.334	& 16.047	& 0.559	& 0.314	& 0.978(0.00)\\
PM$_{2.5}$	& 2.849	& 21.327	& 86.909	& 8.796 & 1.302	& 5.184 & 0.936(0.00)\\
SO2	 & 0.630	& 3.689	& 31.461	& 3.490	& 2.802	& 10.735 & 	0.705(0.00)\\
RH	 & 15.220	& 54.997& 88.766	& 13.425	& -0.341	& -0.485 & 	0.985(0.00)\\
TMP	 & 6.700	& 16.684 & 22.999	& 2.251	&  -0.367	& 0.382 & 	0.992(0.00)\\
WDR	 & 124.183	& 181.283 & 289.713	& 21.723	&  1.103	& 2.131 & 	0.940(0.00)\\
WSP	 & 1.225	& 2.096 & 6.331	& 0.437	&  1.470	& 5.667 & 	0.921(0.00)\\
\bottomrule
\end{tabular}
\caption{Summary statistics of the twelve variables.}
\label{sta}
\end{table}

\section{Methodology}

The data was normalized to guarantee that each feature contributes evenly to the model. For models that are sensitive to feature scales, such as Support Vector Regression and neural networks, this strategy modifies the feature values to have a mean of zero and a standard deviation of one.

The data were divided into four approaches: (1) Considering the variables CO, NO$_X$,  NO, NO$_2$, PM$_{10}$, PM$_{2.5}$, SO$_2$,  O$_3$, RH, TMP, WDR, and WSP. In this scenario we do not add any new feature; (2) Adding temporal features and interactions between the variables;(3) Considering the most important variables using features selection, and (4) Adding temporal lags.

\subsection{Without temporal features and interactions}
For the first approach, the variables used were the following: Concentrations of CO, NO, NO$_2$, NO$_X$, PM$_{10}$, PM$_{2.5}$, SO$_2$, and meteorological variables RH, TMP, WDR, WSP.

\begin{figure}[H]
\centering
\includegraphics[width=0.8\textwidth]{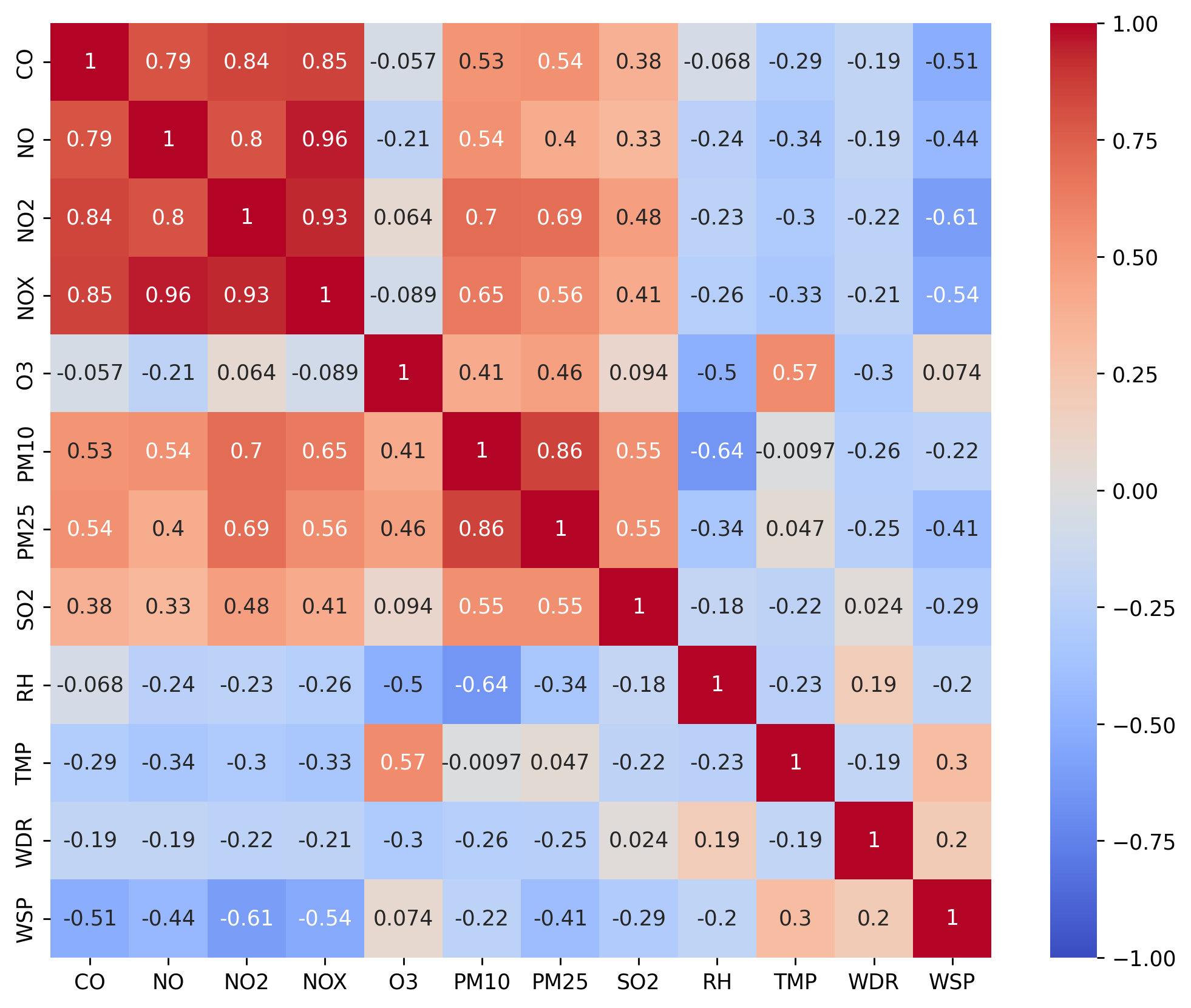}
\caption{Spearman rank correlation matrix between variables.} \label{Fig1}
\end{figure}

We generated a correlation matrix to understand the relationships between different variables in the dataset, Figure  \ref{Fig1}. This graph measures the relationship between two variables in order of their ranks. Therefore, it essentially provides a measure of the monotonic relationship between two variables. The correlation ranges from -1 to 1. If the correlation is close to 1, the features are positively correlated, and if the correlation is close to -1, it means they are negatively correlated. The matrix helps identify which variables have a strong correlation. Feature selection and engineering techniques can be better informed by the matrix. 

\subsection{With temporal features and interactions}

An essential phase in the machine learning process is feature engineering.  It involves the process of transforming raw data into new input features, which can improve the functionality of machine learning models \cite{Hastie,Zeng}. In this study, we used feature engineering to enhance the predictive power of our models.

In this study, we extracted from the dataset temporal features such as the year, month, day, and day of the week. These features may capture seasonal trends and patterns that can influence ozone levels. For instance, ozone levels can vary significantly between weekdays and weekends due to changes in traffic patterns and industrial activity.

In order to capture the interactions between various factors, we also created polynomial features. We seek to model  non-linear interactions between the predictors and the objective variable (ozone levels) by constructing second-degree polynomial terms.

In this approach, we added the temporal features year,  month, day, and day of the week; as well as interaction between the variables. We create columns with cross-products between two features from the original variables. That is, given two variables $x_1$ and $x_2$,  we generate a term $x_1 \times x_2$, but not $x_1^2$ or $x_2^2$.  With this, we introduce non-linear terms into the models that may help to improve their forecasting capacity.

\begin{figure}[H]
\centering
\includegraphics[width=0.8\textwidth]{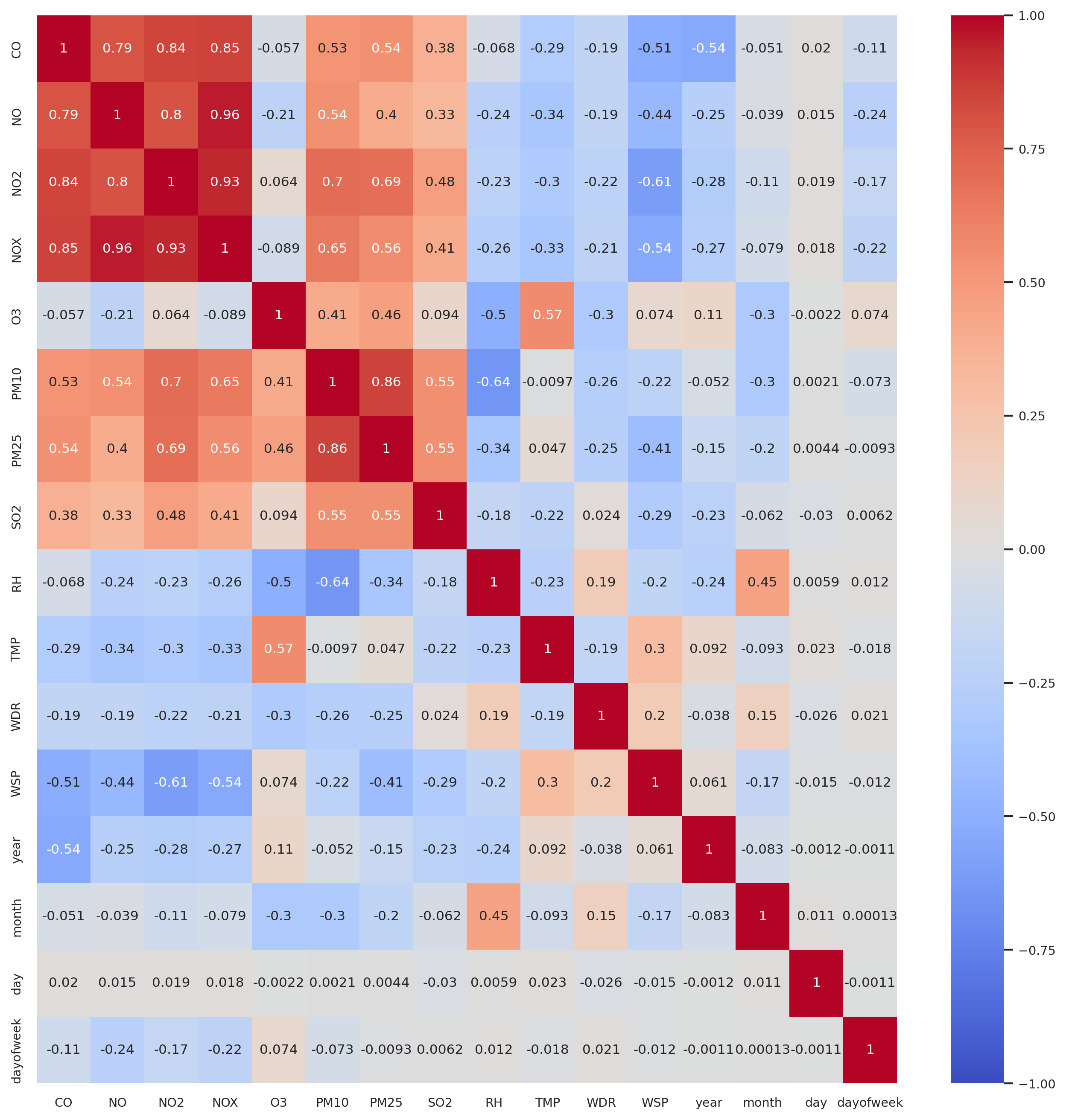}
\caption{Correlation matrix of variables with temporal features}
\label{Fig2}
\end{figure}

Our objective is to improve the models' capacity to predict ozone levels by giving more relevant input variables through feature engineering. The corresponding correlation matrix is shown in Figure \ref{Fig2}, where we show entries with the new temporal features created. Understanding the correlations between variables can help us to create polynomial features and identify potential multicollinearity issues, which can affect model performance.

\subsection{Feature selection}

\begin{figure}[H]
\centering
\includegraphics[width=0.7\textwidth]{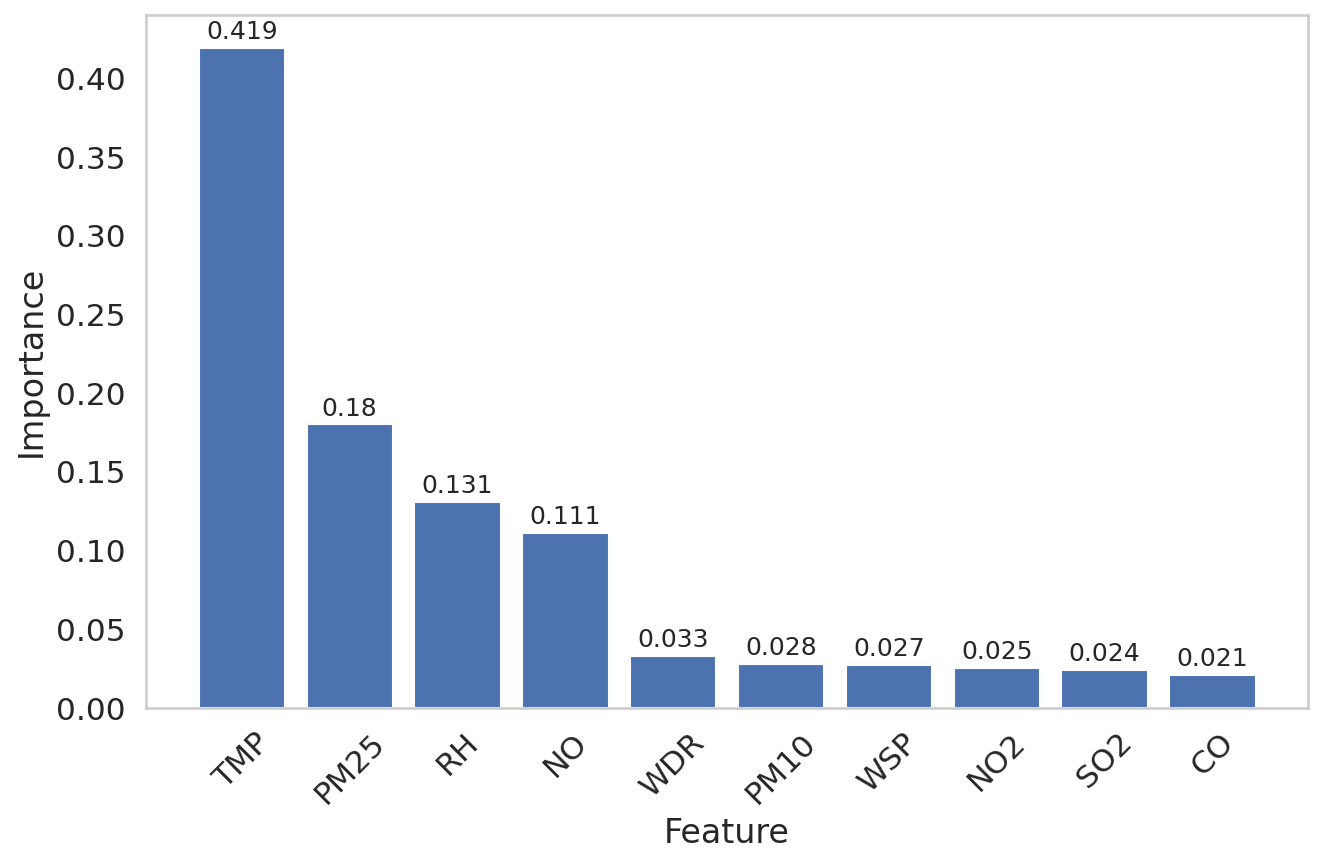}
\caption{Importance of the selected features using the Random Forest model (Pollutants and Meteorological factors).}
\label{caracteristicas}
\end{figure}

 One of the most important steps in creating machine learning models is feature selection. It involves determining the most important features from your dataset that enhance the model's capacity for forecasting. Since Random Forest naturally arranges features based on relevance, it is frequently employed for feature selection. In this section, we use this model to select the most important variables. In  Figure \ref{caracteristicas}, we show the characteristic importance of the variables using Random Forest. To perform the models, we used the variables TMP, PM$_{2.5}$, RH and  NO. 
  
\subsection{With temporal lags}

One popular method for capturing the temporal correlations between observations in sequential data is time lags in time series analysis. For this study, delays have been given to the ozone concentration variable to evaluate the potential impact of past pollutant levels on present concentrations. This technique is especially valuable in environmental and air quality research since pollutant levels often show temporal dependency due to factors such as weather patterns and chemical dynamics in the atmosphere.

In this approach,  we have taken into consideration one- and three-day lags for the O$_3$ concentration variable  The three-day lag covers more extended effects and cyclical patterns, whereas the one-day lag shows the immediate impact of the ozone concentration from the previous day on the present levels.

For this analysis, we considered the variables the relevant columns TMP, PM$_{2.5}$, RH,  NO; as well as the following temporal features year,  month, day, day of the week, 3-day lag, and 1-day lag.

For each  approach, we applied the following regression models:
\begin{itemize}
    \item \textbf{Linear Regression:} A basic linear approach that models the relationship between the dependent variable and one or more independent variables by fitting a linear equation.
    \item \textbf{Random Forest:} An ensemble learning method that constructs multiple decision trees during training and outputs the mode of the classes (classification) or mean prediction (regression) of the individual trees.
    \item \textbf{Gradient Boosting:} An ensemble technique that builds models sequentially, each new model correcting errors made by the previous ones. It optimizes for the best split using a gradient descent algorithm.
    \item \textbf{Support Vector Regression (SVR):} A type of Support Vector Machine (SVM) used for regression tasks, which tries to fit the best line within a threshold value.
    \item \textbf{K-Nearest Neighbors (KNN):} A non-parametric method used for classification and regression, which predicts the value of a point based on the values of its 'k' closest neighbors.
    \item \textbf{ElasticNet:} A linear regression model that combines the properties of both Ridge and Lasso regression, useful for datasets with highly correlated features.
    \item \textbf{XGBoost:} An optimized distributed gradient boosting library designed to be highly efficient, flexible, and portable.
    \item \textbf{LightGBM:} A gradient boosting framework that uses tree-based learning algorithms, known for its efficiency and speed.
    \item \textbf{Bagging Regressor:} An ensemble method that fits multiple versions of a predictor on different sub-samples of the same dataset and averages their predictions.
    \item \textbf{Stacking Regressor:} An ensemble learning technique that combines multiple regression models via a meta-regressor.
    \item \textbf{MLPRegressor (Multilayer Perceptron Regressor):} A neural network model that maps input data to appropriate outputs by learning through backpropagation.
\end{itemize}

\section{Results}

The model results were evaluated using Mean Squared Error (MSE) and the coefficient of determination (R²). Below are the comparative results. Row with MLPRegressor uses a hidden layer with 150 neurons and hyperbolic function as activation function, while the Row MLPRegressor$_1$ uses a hidden layer with 100 neurons with rectified linear unit activation function.

\subsection{Without temporal features and interactions}

\begin{table}[h!]
\centering
\begin{tabular}{lcc}
\hline
\textbf{Modelo} & \textbf{MSE} & \textbf{R\textsuperscript{2}} \\
\hline
Linear Regression & 23.056054 & 0.721882 \\
Random Forest & 19.920742 & 0.759702 \\
Gradient Boosting & 18.652002 & 0.775007 \\
Support Vector Regression & 17.320000 & 0.791074 \\
K-Nearest Neighbors & 24.898125 & 0.699662 \\
ElasticNet & 23.064981 & 0.721774 \\
XGBoost & 18.605508 & 0.775568 \\
LightGBM & 18.717197 & 0.774220 \\
Bagging & 19.912059 & 0.759807 \\
Stacking & 22.270830 & 0.731354 \\
MLPRegressor & 15.563486 & 0.812262 \\
MLPRegressor\_1 & 15.884880 & 0.808386 \\
\hline
\end{tabular}
\caption{Model performance without temporal features and interactions}
\end{table}

\subsection{With temporal features and interactions}

\begin{table}[h!]
\centering
\begin{tabular}{lcc}
\hline
\textbf{Modelo} & \textbf{MSE} & \textbf{R\textsuperscript{2}} \\
\hline
Linear Regression & 19.167475 & 0.768789 \\
Random Forest & 17.639677 & 0.787218 \\
Gradient Boosting & 16.607979 & 0.799663 \\
Support Vector Regression & 16.130527 & 0.805422 \\
K-Nearest Neighbors & 26.187977 & 0.684103 \\
ElasticNet & 20.067817 & 0.757928 \\
XGBoost & 16.396419 & 0.802215 \\
LightGBM & 16.641830 & 0.799255 \\
Bagging & 17.718927 & 0.786262 \\
Stacking & 20.512938 & 0.752559 \\
MLPRegressor & 24.304827 & 0.706818 \\
MLPRegressor\_1 & 15.469999 & 0.813390 \\
\hline
\end{tabular}
\caption{Model performance with temporal features and interactions}
\end{table}

\subsection{Feature selection}

 \begin{table}[h!]
\centering
\begin{tabular}{lcc}
\hline
\textbf{Model} & \textbf{MSE} & \textbf{R\textsuperscript{2}} \\
\hline
Linear Regression & 19.1675 & 0.7688 \\
Random Forest & 17.6397 & 0.7872 \\
Gradient Boosting & 16.6080 & 0.7997 \\
Support Vector Regression & 16.1305 & 0.8054 \\
K-Nearest Neighbors & 26.1880 & 0.6841 \\
ElasticNet & 20.0678 & 0.7579 \\
XGBoost & 16.3964 & 0.8022 \\
LightGBM & 16.6418 & 0.7993 \\
Bagging & 17.7189 & 0.7863 \\
Stacking & 20.5129 & 0.7526 \\
MLPRegressor & 19.0518 & 0.7702 \\
MLPRegressor\_1 & 15.7951 & 0.8095 \\
\hline
\end{tabular}
\caption{Model performance with four characteristics:TMP PM$_{2.5}$, RH,  NO}

\end{table}

\newpage

\subsection{Lags}

 \begin{table}[h!]
\centering
\begin{tabular}{lcc}
\hline
\textbf{Model} & \textbf{MSE} & \textbf{R\textsuperscript{2}} \\
\hline
Linear Regression &	 22.207483 &	0.737668\\
Random Forest	 & 19.167839 &	0.773575\\
Gradient Boosting &	18.011122 &	0.787239\\
Support Vector Regression	 & 18.234035 &	0.784606\\
K-Nearest Neighbors &	25.335282 &	0.700720\\
ElasticNet &	22.207782 &	0.737665\\
XGBoost	 & 18.046307 &	0.786823\\
LightGBM &	17.956106 &	0.787889\\
Bagging &	19.232509 &	0.772811\\
Stacking &	21.905623 &	0.741234\\
MLPRegressor &	17.609166  &	0.791987\\
MLPRegressor\_1	 & 18.128248 &	0.785855\\
\hline
\end{tabular}
\caption{Model performance with lags and following temporal features:TMP PM$_{2.5}$, RH,  NO}

\end{table}

The scatter plots below visualize the actual vs. predicted values for each model,  We only show the plots that compare the models without and with temporal features and interactions (approach 1 and approach 2).

\newpage

\begin{figure}[H]
\centering
\includegraphics[width=0.8\textwidth]{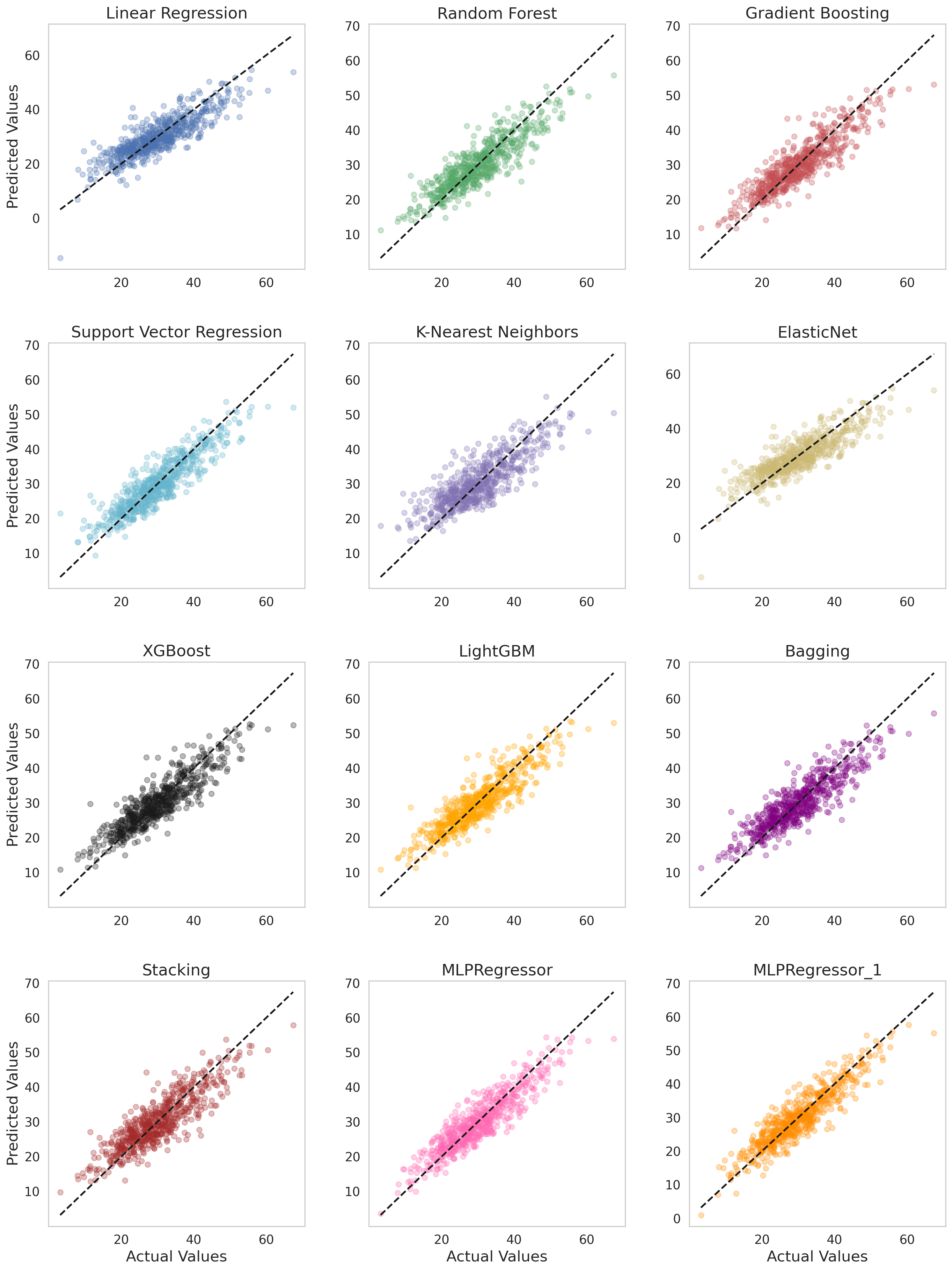}
\caption{Scatter plots \textbf{without} temporal features and interactions (approach 1)}
\end{figure}

\newpage

\begin{figure}[H]
\centering
\includegraphics[width=0.8\textwidth]{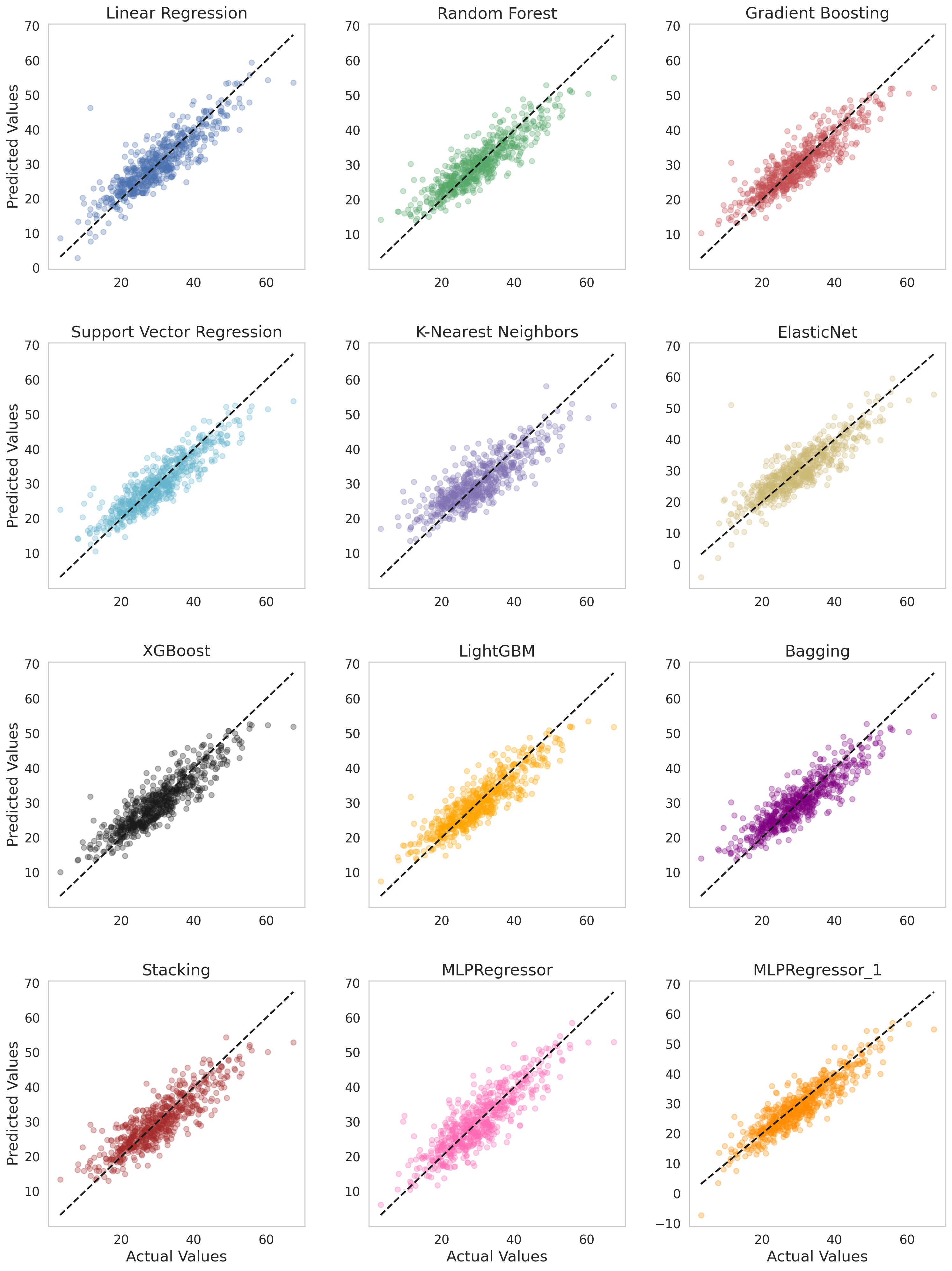}
\caption{Scatter plots \textbf{with} temporal features and interactions(approach 2)}
\end{figure}

\section{Discussion}

The results demonstrate that incorporating temporal features and interactions slightly improves the accuracy of regression models. In most evaluated models, there is a reduction in MSE and an increase in the R² coefficient when temporal features and interactions are added. This suggests that considering the temporal context and interactions between variables provides additional useful information for predicting ozone levels.

The improvement in model performance can be attributed to several factors:

\begin{itemize}
    \item \textbf{Temporal Features:} Adding temporal features such as year, month, day, and day of the week helps capture seasonal variations and periodic trends that affect ozone levels. These features provide the model with a sense of time, which is crucial for understanding and predicting changes in ozone concentrations.
    \item \textbf{Polynomial Features:} Creating polynomial features allows the model to account for non-linear relationships between variables. Ozone formation involves complex chemical reactions influenced by various meteorological conditions and pollutant levels. Polynomial features help the model capture these intricate interactions, leading to better predictions.
    \item \textbf{Enhanced Data Representation:} Feature engineering enhances the representation of the underlying data. By transforming raw data into more informative features, the models can make better use of the available information, resulting in improved predictive accuracy.
    \item \textbf{Reduction of Bias and Variance:} Adding relevant features and interactions can help reduce both bias and variance in the models. Bias reduction occurs because the models can fit the training data more accurately, while variance reduction happens as the models become more robust to different data points by capturing the underlying patterns more effectively.
\end{itemize}

The findings of this study highlight the importance of feature engineering in improving model performance for predicting ozone levels. By considering temporal features and interactions, we can develop more accurate models that better understand the dynamics of ozone formation.

\section{Conclusion}

Accurate prediction of ozone levels is crucial for protecting public health and the environment. This study shows that incorporating temporal features and interactions between variables improves the accuracy of regression models. By adding features that capture temporal trends and interactions, we can provide models with more informative input variables, enhancing their ability to predict ozone levels accurately.

Therefore, it is recommended to consider these features in future models for predicting ozone and other atmospheric pollutants. The findings of this study can be applied to other pollutants and environmental monitoring tasks, highlighting the importance of feature engineering in machine learning.

\vspace{1cm}

\textbf{Authors declared no conflict of interest} 

\medskip

\textbf{We have not received any funding for this research} 

\medskip

\textbf{Generative AI tools were not used for the writing of the manuscript}

\end{document}